\documentclass[conference]{IEEEtran}
\IEEEoverridecommandlockouts
\usepackage[numbers]{natbib}
\usepackage[textwidth=3.5cm,textsize=scriptsize]{todonotes}
\usepackage{amsmath,amssymb,amsfonts}
\usepackage{algorithmic}
\usepackage{graphicx}
\usepackage{textcomp}
\usepackage{colortbl}
\usepackage{multirow}
\usepackage{xcolor}
\usepackage{subcaption}
\usepackage{rotating}
\usepackage[inline]{enumitem}

\newtheorem{definition}{Definition}

\newcommand{\ind}[1]{\ensuremath{\mathit{#1}}}
\newcommand{\con}[1]{\ensuremath{\mathit{#1}}}
\newcommand{\var}[1]{\ensuremath{#1}}
\newcommand{\fun}[1]{\ensuremath{\mathit{#1}}}
\newcommand{\func}[2]{\ensuremath{\fun{#1}\left(#2\right)}}
\newcommand{\matrice}[1]{\ensuremath{#1}}

\newcommand{\x}[1]{\ensuremath{\var{x}_{#1}}}

\newcommand{\X}[1]{\ensuremath{\matrice{X}_{#1}}}

\newcommand{\Z}[1]{\ensuremath{\matrice{z}_{#1}}}

\let\oldTheta\theta
\renewcommand{\theta}[1]{\ensuremath{\matrice{\oldTheta}_{#1}}}

\let\oldMu\mu
\renewcommand{\mu}{\ensuremath{\oldMu\left(\X{};\theta{1}\right)}}
\let\oldSigma\Sigma
\renewcommand{\Sigma}{\ensuremath{\oldSigma\left(\X{};\theta{1}\right)}}

\newcommand{\p}[2]{\ensuremath{\fun{P}_{#1}\left(#2\right)}}

\let\oldProd\prod
\renewcommand{\prod}[2]{\ensuremath{\oldProd_{#1=1}^{#2}}}

\let\oldSum\sum
\renewcommand{\sum}[2]{\ensuremath{\oldSum_{#1=1}^{#2}}}

\let\oldInt\int
\renewcommand{\int}[2]{\ensuremath{\oldInt_{#1=1}^{#2}}}

\makeatletter
\newcommand\xleftrightarrow[2][]{%
  \ext@arrow 9999{\longleftrightarrowfill@}{#1}{#2}}
\newcommand\longleftrightarrowfill@{%
  \arrowfill@\leftarrow\relbar\rightarrow}
\makeatother

\definecolor{colorblue}{rgb}{0.29, 0.59, 0.82}
\definecolor{colorred}{rgb}{0.9, 0.17, 0.31}
\definecolor{colorred}{rgb}{0.9, 0.17, 0.31}
\definecolor{coolgrey}{rgb}{0.55, 0.57, 0.67}
\newcommand{\dcb}{\cellcolor{colorblue!20}}

\newcommand{\dcrh}{\rowcolor{coolgrey!30}}
\newcommand{\dcrf}{\rowcolor{coolgrey!20}}

\newcommand{\mr}[1]{\multirow{2}{*}{#1}}
\newcommand{\mrm}[1]{\multirow{-2}{*}{#1}}

\newcolumntype{"}{@{\hskip\tabcolsep\vrule width 1.5pt\hskip\tabcolsep}}
\makeatletter
\newcommand{\thickhline}{%
    \noalign {\ifnum 0=`}\fi \hrule height 1pt
    \futurelet \reserved@a \@xhline
}

\def\BibTeX{{\rm B\kern-.05em{\sc i\kern-.025em b}\kern-.08em
    T\kern-.1667em\lower.7ex\hbox{E}\kern-.125emX}}
\begin{document}

\title{Conditional Constrained Graph Variational Autoencoders for Molecule Design\\
}

\author{
\IEEEauthorblockN{Davide Rigoni}
\IEEEauthorblockA{\textit{Department of Mathematics} \\
University of Padua, Padua, Italy \\
\textit{Fondazione Bruno Kessler} \\
Via Sommarive, 18, Povo, Italy \\
davide.rigoni.2@phd.unipd.it \\
}
\and
\IEEEauthorblockN{Nicolò Navarin, member, IEEE}
\IEEEauthorblockA{\textit{Department of Mathematics} \\
University of Padua, Padua, Italy \\
nnavarin@math.unipd.it}
\and
\IEEEauthorblockN{Alessandro Sperduti, senior member, IEEE}
\IEEEauthorblockA{\textit{Department of Mathematics} \\
University of Padua, Padua, Italy \\
sperduti@math.unipd.it}
}


\maketitle
\begin{abstract}
In recent years, deep generative models for graphs have been used to generate new molecules.
These models have produced good results, leading to several proposals in the literature.
However, these models may have troubles learning some of the complex laws governing the chemical world.
In this work, we explore the usage of the histogram of atom valences to drive the generation of molecules in such models.
We present Conditional Constrained Graph Variational Autoencoder (CCGVAE), a model that implements this key-idea in a state-of-the-art model, and shows improved results on several evaluation metrics on two commonly adopted datasets for molecule generation.

\end{abstract}


\section{Introduction}
\label{sec:background}
The search for new molecules capable of exhibiting specific target properties is a long-standing problem in chemistry.
New molecules could improve technologies in many industrial and pharmaceutical areas. 
For example, new molecules with specific target properties can become new drugs that can improve treatments for diseases or even cure new diseases\footnote{https://www.covid19.jedi.group/}.
Many approaches are used to explore the chemical space such as high throughput screening~\cite{curtarolo2013high, pyzer2015high} and evolutionary algorithms~\cite{devi2015evolutionary}.
Alternative approaches use Machine Learning models to predict properties of pre-defined or commercial compounds, e.g. using Recursive Neural Networks \cite{QSPRAR_NNS, DBLP:journals/jcisd/BernazzaniDMMSST06}. 
In recent years the machine learning community has devoted much effort in the study of machine/deep learning models that are capable of generating candidate molecules that are likely to exhibit some pre-specified properties~\cite{Oglic2018}.
Thanks to the development of increasingly effective deep learning models and the presence of large datasets, this approach achieved promising results.

In literature, there are many proposals for generative models for molecules, that are often applied to the drug generation domain. 
The majority of them are based on the following approaches: 
\begin{enumerate*}
    \item variational autoencoders~\cite{diederik2014auto};
    \item generative adversarial networks~\cite{arjovsky2017wasserstein};
    \item recurrent neural networks~\cite{segler2018generating, preuer2018frechet};
    \item adversarial autoencoders~\cite{makhzani2015adversarial}.
\end{enumerate*}
To be able to compare all these different families of approaches, some frameworks have been developed as baselines, such us MOSES~\cite{polykovskiy2018molecular}, and GuacaMol~\cite{brown2019guacamol}.

In this work, we focus on the variational autoencoder approach, which seems to deliver the best trade-off between generative capabilities and ease of training. In fact, adversarial based models are quite problematic to train, while  models based on recurrent neural networks make it difficult to control the ``shape'' of the latent space and thus its sampling for the generation of novel compounds.

Early works adopting the \emph{variational autoencoder} approach~\cite{gomez2018automatic, DBLP:conf/icml/KusnerPH17, DBLP:conf/iclr/DaiTDSS18} share the
capability of generating new molecules using SMILES~\cite{weininger1988smiles, weininger1989smiles, weininger1990smiles} representations, which are strings describing the structure of the molecules.
In particular, Grammar VAE~\cite{DBLP:conf/icml/KusnerPH17} adds a \emph{context-free grammar} to the standard \emph{variational autoencoder} \cite{gomez2018automatic} to guide the correct generation of SMILE strings, while Syntax Directed VAE~\cite{DBLP:conf/iclr/DaiTDSS18} adds a more expressive grammar, the \emph{attribute grammar}~\cite{knuth1968semantics}, capable of generating a higher number of valid SMILES strings.
Other models propose to directly generate a graph representation of the molecule~\cite{DBLP:conf/icml/JinBJ18, DBLP:conf/nips/MaCX18, DBLP:conf/nips/LiuABG18}.
Specifically,
Junction Tree VAE~\cite{DBLP:conf/icml/JinBJ18} represents molecules using graphs composed of chemical substructures that are extracted from the training set. New molecular graphs are obtained by first generating a tree-structured scaffold formed by substructures (the junction tree), and then combining the substructures together using a graph message passing network.
Regularized Graph VAE~\cite{DBLP:conf/nips/MaCX18} transforms the constrained molecule generation problem into a regularized unconstrained problem, applying a generalization of the Lagrangian function in order to move all the constrains into the optimization function.
State-of-the-art Constrained Graph VAE (CGVAE)~\cite{DBLP:conf/nips/LiuABG18} uses a \emph{variational autoencoder} to generate a latent space normal distribution from which the model samples a point for each atom, differently from previously models that sample only one point per molecule.

A common and crucial feature of the above models is to embed chemical background knowledge, e.g.  to increase  the number of generated valid molecules Syntax Directed VAE exploits a chemical-based \emph{context-free grammar},  while CGVAE  uses  the valence of each atom  to guide the generation of the bonds.

Although CGVAE returns state-of-the-art results, we argue that its choice to sample the latent space independently for each atom of the generated molecule, constitutes a limitation of the model.
In fact, such independence assumption, due to the chemical/physical laws that govern the chemical realm, is clearly not satisfied in general. Thus, introducing a form of dependence among atoms, should help the model to better learn the distribution of the molecular structures in input. 
In order to remove this independence assumption, in this paper we present the Conditional Constrained Graph VAE (CCGVAE) model, in which \emph{histograms of valences} of the molecules are used to make the generation of atoms from the latent space dependent on the already sampled atoms.
Specifically, given a valid target valence histogram for the molecule to be generated, the atom generation process is designed so to keep the current valence histogram obtained by the already generated atoms to stay compatible with the target histogram.
Thus, our model is a variant of the CGVAE model, in which the decoder is {\em enhanced} to incorporate the histograms of the valences.

This paper is organized as follows.
In Section~\ref{sec:cgvae} we provide the background information about the CGVAE model.
Section~\ref{sec:model} presents our proposal for the generation of atoms driven by the histogram of valences.
In Section~\ref{sec:results}, we discuss the experimental evaluation of the model.
Section~\ref{sec:conclusions} concludes the paper.
The code used for this work is published online\footnote{https://github.com/drigoni/ConditionalCGVAE}.

\section{Constrained Graph VAE}
\label{sec:cgvae}
Let us start with some definitions.
In this paper, we consider molecules represented as graphs. A graph is a tuple $G=(V,E,\mathcal{L})$ where $V=\{v_1,\ldots,v_m\}$ is the set of nodes representing atoms, $E$ is the set of edges representing bonds between atoms, and $\mathcal{L}$ is a function associating labels to nodes (the atom type) and edges (the bond type).

Constrained Graph Variational Autoencoder (CGVAE) is a model proposed by~\citeauthor{DBLP:conf/nips/LiuABG18}~\cite{DBLP:conf/nips/LiuABG18}.
This model is based on the variational autoencoder approach in which there are two main components: the \emph{encoder} and the \emph{decoder}.
Given in input a molecule, the \emph{encoder} learns to encode its atoms in a latent space defined by a normal distribution \( \mathcal{N}\left(0, I\right)\), where $I$ is the identity matrix. 
Since the latent space is defined by a standard normal distribution, it is possible to sample new points, i.e. atoms, from that distribution and 
to decode them into a valid molecule by the \emph{decoder}. The decoder does that by incrementally adding bonds among the sampled atoms.
Specifically, at the beginning the model randomly selects an initial atom (focus) and adds (valid) bonds with other atoms till a stop criterion is satisfied. Whenever a new bound with an unconnected atom is inserted, the involved atom is inserted into a queue. When the focus atom reaches the stop criterion, an atom is extracted from the queue and it becomes the new focus atom. The generation of new bounds is then resumed with the new focus atom. It should be noticed that, the generation of a new (valid) bound (and its type) at a specific decoding point is conditioned on the current partial molecule built till that point. The decoding process stops when the queue is empty. Finally, eventually isolated atoms are removed, and the obtained molecule is returned as output of the decoder. \\
In addition to that, the model incorporates an \emph{optimization} component which allows to drive the generation process towards molecules that exhibit a high pre-specified property. This is achieved starting from random points in the latent space (i.e., set of atoms) and performing gradient ascent/descent in latent space with respect to the property of interest.

In the following, we summarize the main computational steps of the model, i.e. encoding, decoding, optimization, and training.

\subsection{Encoding}
\label{subsec:cgvae_encoder}
The first computational step of the CGVAE model is constituted by the \emph{encoder}.
At the beginning the encoder receives in input the graph representation of a molecule with $|V|=m$
atoms.
Then each atom \var{v} $\in V$ is encoded in a normal probability distribution parameterized by mean \(\oldMu_v\) and variance \(\oldSigma_v\).
Specifically, a Gated Graph Recurrent Neural Network (GGRNN)~\cite{DBLP:journals/corr/LiTBZ15} with residual connections is first used to devise a hidden representation of each atom (node) that embeds  the information of its neighbors.
Then, the encoder uses two neural networks\footnote{Both the neural networks preserve the input dimension and are built using only one layer with linear activation function. The dimension used is fixed to 100.} in order to generate means and variances for each atom from the last updated hidden state values. 
The vector \Z{v} representing the atom \var{v} encoded in the latent space is then sampled according to the distribution \( \mathcal{N}\left(\oldMu_v,\oldSigma_v\right)\).
It should be emphasized that this model generates a probability distribution for each node in the graph, unlike other approaches~\cite{gomez2018automatic, DBLP:conf/icml/KusnerPH17, DBLP:conf/iclr/DaiTDSS18, DBLP:conf/icml/JinBJ18, DBLP:conf/nips/MaCX18, de2018molgan} that generate a probability distribution per molecule.
It follows that while in the other models the sampling of a point in the latent space implies the choice of a precise molecule, in this case one sampling in the latent space implies only the choice of a single atom, making it necessary to use more samples in the decoder to perform the generation of a new molecule.

\subsection{Decoding}
\label{subsec:cgvae_decoder}
The decoder receives as input a set of vectors $\{\Z{v}\}_{v \in [1,m]}$, where each \Z{v} represents an atom.
During the generation phase, \Z{v}'s are sampled from the normal distribution \(\mathcal{N}\left(0, \con{I}\right)\), where \con{I} is the identity matrix.
During training, the values \Z{v} are obtained by the distribution \( \mathcal{N}\left(\oldMu_v,\oldSigma_v\right)\), using the \emph{reparameterization trick}.
The nodes are initially not connected with each other and they are associated with a state \(\var{h}_v^{t=0} = [\Z{v}, \tau_v]\) in which \(\tau_v\) is
supposed to be a one-hot vector indicating the node type, obtained by sampling from a learned softmax transformation from the latent space to the atom type space, i.e. $\tau_v \sim softmax(f(\Z{v}))$. Actually, in the software implementing the model it is a 100-dimensional vector embedding of the node type.

Once all the atoms hidden states are obtained, a vector \(\var{H}_{init}\) considering global information conveyed by all initially disconnected nodes is computed at step \ind{t = 0} by averaging the vectors $\{h_v^0\}_{v \in V}$. Subsequently, at each step \ind{t} the vector \(\var{H}^t\) is computed by considering only connected nodes, i.e., excluding isolated nodes.
The generation of bonds  proceeds as described at the beginning of the section.

Whenever a new bond is added and consequently the partially constructed graph is modified, the model uses a Gated Graph Recurrent Neural Network (GGRNN) to update the information of each node, generating new states \(\var{h}_v^{t+1}\) for each step $s \in [0, 12]$:
\begin{displaymath}
\var{m}_v^0=\var{h}_v^0 ,\ \var{m}_v^{s+1}=GRU\left[\var{m}_v^s, \oldSum_{v\xleftrightarrow{l}u}E_l\left(\var{m}_u^s\right)\right]  ,\ \var{h}_v^{t+1}=\var{m}_v^S
\end{displaymath}
where $h_v^t$ is the hidden representation of the atom $v$ at the time step $t$ and the summation is over all the neighbors of node $v$ connected by an arc of type $l$ for each value of $l$.
\(E_l\) is a neural network for edges of type $l$ that preserves the dimension.

In order to decide which specific bound (and its type $l$) to add, 
the following distribution over candidate edges is used:
\begin{displaymath}
\p{}{v\xleftrightarrow{l}u|\var{\phi}_{v,u}^t}\!\! = \! \p{}{v\xleftrightarrow{l}u|\var{\phi}_{v,u}^t, v\xleftrightarrow{}u} \cdot \p{}{v\xleftrightarrow{}u|\var{\phi}_{v,u}^t}
\end{displaymath}
where
\(
\var{\phi}_{v,u}^t = 
[\var{h}_v^t, \var{h}_u^t, \func{d}{v,u}, \var{H}_{init}, \var{H}^t
],\)
\func{d}{v,u} is the distance calculated between nodes \ind{v} and \ind{u} over the graph currently generated, $v\xleftrightarrow{}u$ indicates the presence of a bond between the atoms $v$ and $u$, and $v\xleftrightarrow{l}u$ indicates the presence of a bond of type $l$ between $v$ and $u$.
With the vector \(\var{\phi}_{v,u}^t\) that groups the local information of the nodes and the global information on the graph, the model uses the function \func{C}{\phi} to obtain the probability of existence of the edge, i.e. the bond, between the two atoms \var{v} and \var{u}. 
Moreover it uses the function \func{L_l}{\phi} to obtain the edge's type probabilities.
\func{C}{\phi} and \func{L_l}{\phi} are fully connected networks with a single hidden layer of 250 units and \emph{ReLU} non-linearities.
During the generation of new molecules, the existence and the type of the bonds are sampled from the probabilities returned by the two functions, while during training \emph{teacher forcing} is used.
The probabilities are defined as:
\begin{displaymath}
\p{}{v\xleftrightarrow{}u|\var{\phi}_{v,u}^t} = \dfrac{\var{M}_{v\xleftrightarrow{}u}^t exp\left[\func{C}{\phi_{v,u}^t} \right]}{\oldSum_w\var{M}_{v\xleftrightarrow{}w}^t exp\left[\func{C}{\phi_{v,w}^t} \right]}, \\
\end{displaymath}
\begin{displaymath}
\p{}{v\xleftrightarrow{l}u|\var{\phi}_{v,u}^t, v\xleftrightarrow{}u} = \dfrac{\var{m}_{v\xleftrightarrow{l}u}^t exp\left[\func{L_l}{\phi_{v,u}^t} \right]}{\oldSum_k\var{m}_{v\xleftrightarrow{k}w}^t exp\left[\func{L_k}{\phi_{v,w}^t} \right]}, \\
\end{displaymath}
where both \func{C}{\phi} and \func{L_l}{\phi} are represented by neural networks in which the last layer has a \emph{softmax} function on the probabilities of outputs weighted with the use of binary masks \(\var{M}_{v\xleftrightarrow{}u}^t\) and \(\var{m}_{v\xleftrightarrow{l}u}^t\). Similarly to those used in~\cite{DBLP:conf/icml/KusnerPH17}, they are intended to prohibit bonds that violate certain chemical constraints in the construction of molecules.

As already said at the beginning of the section, the generation of bonds connected to the focus atom at time step $t$ stops when a given criterion is met. The criterion is met when a connection with a special \emph{stop-node} \(\oslash\) is created.
In this case 
the generation process proceeds by selecting a new node (i.e., the new focus atom) on which to apply the procedure again.
Every time a new atom is connected, it is pushed in a \emph{FIFO structure} which is queried to obtain the next node to visit. Thus, the model visits graph nodes using a breath first algorithm.
At the end of the process, when the \emph{FIFO structure} is empty i.e. all the (connected) nodes have been visited, all the bonds necessary for the validity of the molecule are completed by adding hydrogen atoms bonded with all the atoms whose valences are not correct. A public online software was used to manage the molecular structure\footnote{RDKit: Open-source cheminformatics; http://www.rdkit.org}.

\subsection{Optimization}
\label{subsec:cgvae_optimization}
CGVAE directly incorporates in the \emph{variational autoencoder} the neural network \fun{O} which has the purpose of optimizing the latent space towards molecules that exhibit better values for a target property.
Thanks to \fun{O} it is possible to use gradient ascent/descent\footnote{Depending if the target property should be maximized or minimized.} on the latent space to find a new set of latent points that is decoded into a molecule with a better value for the target property. 

\subsection{Training}
\label{subsec:cgvae_training}
The decoder is trained by supervising the creation of the arcs through breath first search. Thus, a data pre-processing step is needed.
Since the idea of this model is to generate the edges from a set of initially disconnected nodes, 
the model should compute the marginalized probability of each edge with respect to all possible node permutations. Since this would be computationally expensive, Monte Carlo sampling is used to get an estimation.
The loss function to minimize is:
\begin{displaymath}
    \mathcal{L}=\mathcal{L}_{recon}+\lambda_1\mathcal{L}_{latent}+\lambda_2\mathcal{L}_{opt}
\end{displaymath}
where \(\mathcal{L}_{recon}\) is the decoder loss and \(\mathcal{L}_{latent}\) is the \emph{variational autoencoder} Kullback–Leibler loss and $\mathcal{L}_{opt}$ is the optimization network loss.
\(\lambda_1\) and \(\lambda_2\) are two constant scalar values set to $0.3$ and $10$, respectively.
In particular, the \(\mathcal{L}_{recon}\) loss is calculated as the sum of the \emph{cross-entropy} loss calculated on the predicted atom types probabilities with respect to the real atom types, and the \emph{cross-entropy} loss calculated on the predicted bonds probabilities  with regard to the real bonds at each time step.
\(\mathcal{L}_{opt}\) loss is calculated as the \emph{Mean Squared Error} between the predicted and the target property values. 
\section{Proposed Model: Conditional CGVAE}
\label{sec:model}

\begin{figure}[t]
\includegraphics[width=\linewidth]{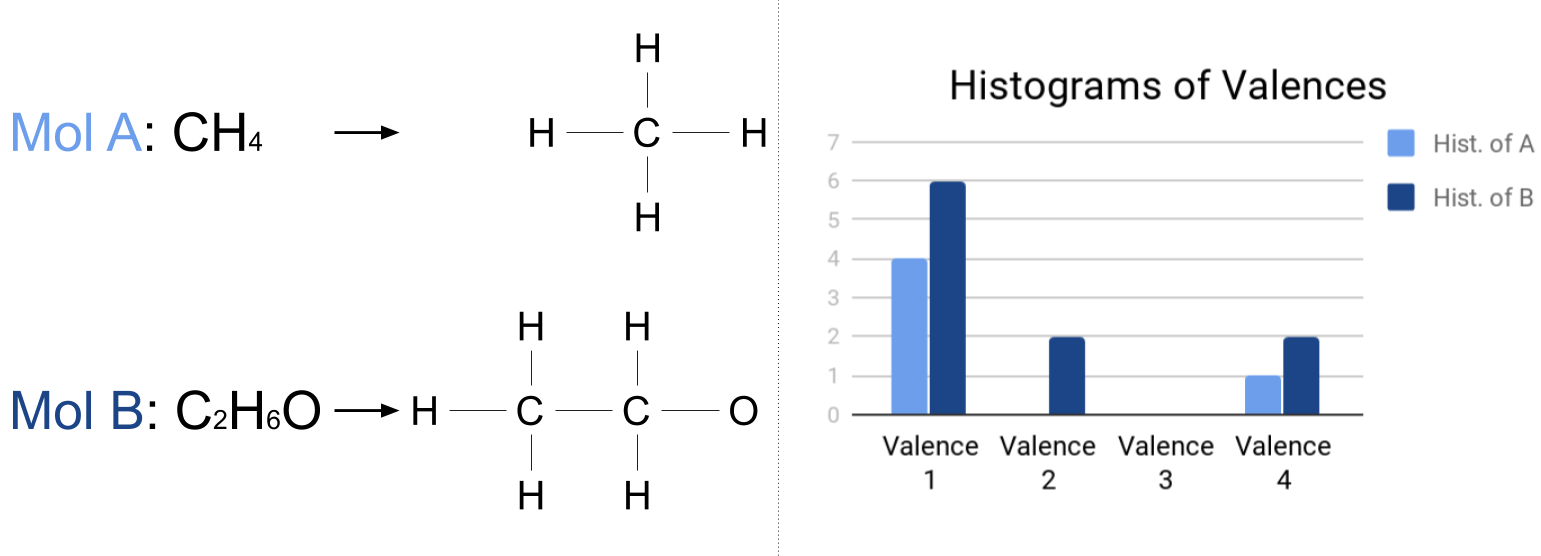}
\caption{Histograms compatibility example. In the left part of the figure we can see the molecule \textbf{A} (Methane) and the  molecule \textbf{B} (Ethanol). In the right part of the figure we report the corresponding histograms of the two molecules.  By recalling that \textbf{C}arbon has valence 4, \textbf{O}xygen has valence 2 and \textbf{H}ydrogen has valence 1, it can be noticed that the histogram of  molecule \textbf{A} is compatible with that one of  molecule \textbf{B}.}
\label{fig:hist_example}
\end{figure}

This work is based on the CGVAE model (Section~\ref{sec:cgvae}).
We started from the Constrained Graph VAE model code made available\footnote{https://github.com/microsoft/constrained-graph-variational-autoencoder} by the authors, in which 
we enhanced the \emph{decoder}.
Let us start providing some definitions.
\begin{definition}[\textbf{Histogram of valences}]
Given a molecule with $m$ atoms and maximum atom valence $\nu \in \mathbb{N}$, the histogram of valences $\alpha$ is the histogram built considering all the valences of the $m$ atoms, where $\alpha[i]$ with $i \in \{1, .., \nu\}$ is the number of atoms with valence equal to $i$.
\end{definition}
\begin{definition}[\textbf{Histograms distribution}]
Given a dataset of molecules $Tr$ in which each molecule is associated with its histogram of valences, the histograms distribution ${\cal{H}}_{Tr}$ is the probability distribution obtained considering all the histograms of valences of molecules in $Tr$.
\end{definition}
\begin{definition}[\textbf{Histogram compatibility}]
Let $\alpha$ and $\beta$ be two histograms of valences belonging to two possibly different molecules, in which $\nu \in \mathbb{N}$ is the maximum atom valence value for both molecules.
We say that $\alpha$ is compatible with $\beta$:
\begin{displaymath}
    iff \quad \forall i \in \{1, .., \nu\}, \quad \beta[i] \geq \alpha[i]
\end{displaymath}
\end{definition}
In other words, histogram $\alpha$ is compatible with a second histogram $\beta$ if and only if, for each valence, the value present in $\beta$ is greater than or equal to that in $\alpha$.
Figure~\ref{fig:hist_example} shows an histograms compatibility example. 

The basic idea is to exploit the histogram of valences to drive the atom type assignment process. Specifically, at the beginning of the generation process, in addition to the number of atoms $|V|=m$, we provide a reference histogram $\alpha_0$ with total count equal to $m$, selected at random from the ones computed from molecules in the training set. The assignment of an atom type to each point sampled from the latent space is then conditioned to the already assigned atom types.
As data pre-processing, the model associates to each molecule its \emph{histogram of valences} and calculates the \emph{histograms distribution} ${\cal H}$ considering all the molecules in the training set.
Given a molecule in input to the model, we use its histogram of valences and the histograms distribution ${\cal H}_{Tr}$ to condition the decoder during the reconstruction phase.
In particular, in order to condition the decoder with the histogram of valences information, we modify the initial part of the decoder which assigns the atom type to each sample from the latent space, so to make it dependent on the atom type assignment history. We do that by introducing a new recurrent component at the first part of the decoder.

\begin{figure*}[t]
\includegraphics[width=\linewidth]{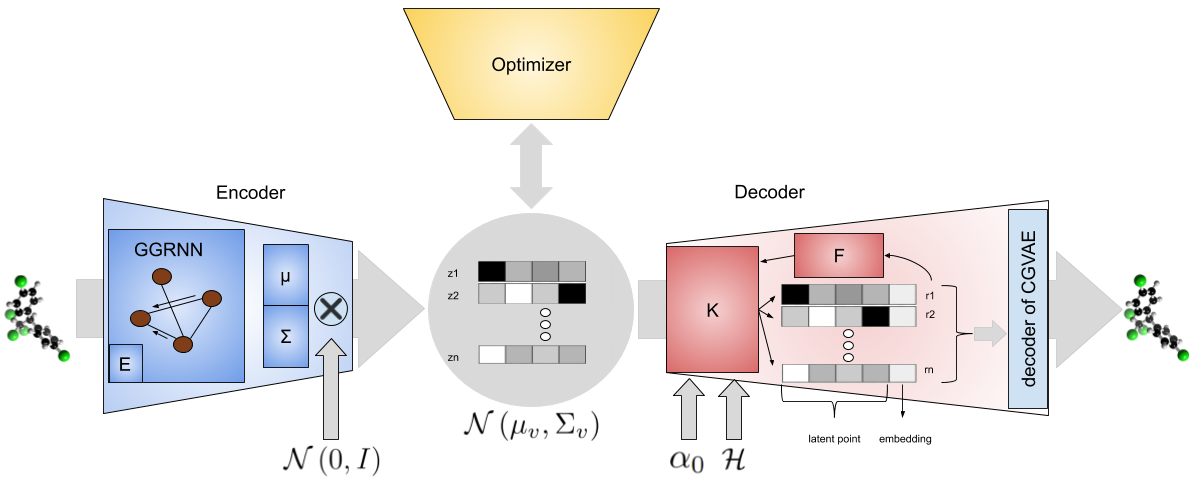}
\caption{Conditional Constrained Graph Variational Autoencoder model structure (CCGVAE). The molecule in input to the encoder goes through the GGRNN with the edge-specific neural network \fun{E} and then the encoder encodes the molecule in the latent space \var{Z}. The first part of the decoder receives in input the sampled points, the initial histogram $\alpha_0$ and the \emph{histograms distribution} ${\cal H}$. $K$ is a function that generates the embedding part for each atom, while $F$ is a function that generates the atom types probabilities.
}
\label{fig:ccgvae}
\end{figure*}

Figure~\ref{fig:ccgvae} provides an overview of the CCGVAE structure, where
the new decoder is represented by two distinct phases which are placed in succession one after the other. 
The first phase generates nodes, while the second phase generates bonds. While the second phase stays the same as defined in CGVAE, the first phase is modified as described in the following.

\subsection{Conditional Atom Type Assignment}
Here we describe how the valence histograms are used to condition the atom type assignment to samples from the latent space.
The decoder receives as input the $m$ latent space points \Z{v} with $v \in \{1, .., m\}$, the histograms distribution ${\cal H}_{Tr}$ and a histogram $\alpha_0$ (of the considered molecule in reconstruction during training, or randomly sampled from ${\cal H}_{Tr}$ in generation).
\Z{v} is sampled from the distribution \(\mathcal{N}\left(\oldMu_v, \oldSigma_v \right)\) during the learning procedure and from  \(\mathcal{N}\left(0, I\right)\) during the generation procedure.
Let $\alpha^u_0$ the histogram where all the valences are 0 and $t \in \{1, .., m\}$, then each atom type is predicted using the following procedure:
\begin{align*}
    \alpha^d_{t}             &= \alpha_{t-1} - \alpha^u_{t-1}, \\
    \var{R}_{t}&= K(\Z{t}, \alpha^d_{t}, \alpha^u_{t-1}), \\
    \tau_{t}            &= \text{Sample}_\text{type}(F(\var{R}_{t}), \alpha^d_{t})\\
    \alpha^u_{t}           & = \text{Update}(\tau_{t}, \alpha^u_{t-1}),\\
    \alpha_{t}               &= \text{Sample}_\text{distr}({\cal H}, \alpha^u_t),
\end{align*}
where:
\begin{itemize}
	\item $K(\Z{t}, \alpha^d_t, \alpha^u_t)$ is a function that receives in input $\Z{t}$, the difference histogram $\alpha^d_t$ and the updated histogram $\alpha^u_t$ at each step $t$.
    This function maps the input to a new representation $\var{e}_t$ of the atom $t$ according to the two histograms in input. In the end, this function returns an atom hidden representation $\var{R}_t =\left[\Z{t},\var{e}_t \right]$;
	\item $F(\var{R}_t)$ is a function that receives in input an atom hidden representation $\var{R}_t$, and generates a probability distribution on the atom types;
	\item $\text{Sample}_\text{type}(F(\var{R}_t), \alpha^d_t)$ is a function that samples the atom type from the distribution returned by the function \fun{F} 
	applying a binary mask in order to remove all the atoms whose valences have a zero-value in the histogram $\alpha^d_t$. At training time \emph{teacher forcing} is used, while at generation time and at reconstruction time, the binary mask is used in order to improve the accuracy;
	\item $\text{Update}(\tau_t, \alpha^u_t)$ is a function that updates the histogram $\alpha^u_t$ with the valence of the sampled atom type $\tau_t$;
	\item $\text{Sample}_\text{distr}({\cal H}, \alpha^u_t)$ is a function that, at training time and during the reconstruction task, returns always the histogram $\alpha^v_0$. During the generation of new molecules this function samples from ${\cal H}$ a new histogram $\alpha_{t+1}$ with at least $m$ atoms, such that $\alpha^u_t$ is compatible with $\alpha_{t+1}$
	If there is no compatible histogram in ${\cal H}$, a random sampling is done according to ${\cal H}$ with no further constraints.
\end{itemize}
The benefit of the above procedure consists in avoiding the generation of a set of atoms that 
is not compatible with the histograms of valences of the molecules in the training set, in principle constraining learning and
generation to focus on the original generating distribution of the training set.

\subsection{Bonds Generation}
This part of the model receives as input the set of nodes generated by the previous phase and, using the same process as CGVAE, starting from a node it proceeds with the generation of the bonds towards the other atoms until the molecule is completed.

\section{Experiments}
\label{sec:results}
Following~\cite{rigoni2020assessment}, we 
compared our model with several state-of-the-art proposals
on two datasets, using different metrics in order to see the potential of each model. In particular, for each model we assessed the ability to reconstruct the input molecules and the ability to generate new ones.


\subsection{Datasets and Metrics}
\label{subsec:data_metrics}
We consider two datasets of molecules: QM9~\cite{ruddigkeit2012enumeration,ramakrishnan2014quantum}, composed by about 134,000 organic molecules with a maximum of 9 atoms, and  ZINC~\cite{irwin2005zinc}, composed by 250,000 drug-like molecules with up to 38 atoms.
More details on the dataset molecules are reported in Table~\ref{tab:datasets}.
We use the same training and test splits in each dataset for each model.
However, differently from~\cite{rigoni2020assessment}, we have chosen a different split of the data in the QM9 dataset as the test set in the original split did not reflect the distribution of molecules present in the training set very well.
For this reason, we re-evaluated all the baseline models on the QM9 dataset. Thus, the results we report are slightly different from the ones in literature.
\begin{table}[t]
\begin{center}
    \begin{tabular}{|c|c|c|c|c|}
        \hline
        \dcrh \textbf{dataset} & \textbf{\#Molecules}& \textbf{\#Atoms}& \textbf{\#Atom Types} & \textbf{\#Bond Types}\\
        \hline
        QM9     & 134K      & 9         & 4     & 3     \\
        \hline
        ZINC    & 250K      & 38        & 9     & 3     \\
        \hline
    \end{tabular}
    \label{tab:datasets}
\end{center}
\caption{Statistics of the QM9 and ZINC datasets.}
\label{tab:datasets}
\end{table}

\begin{figure}[t]
\centering
\begin{subfigure}[t]{0.90\linewidth}
\centering
\includegraphics[width=\linewidth]{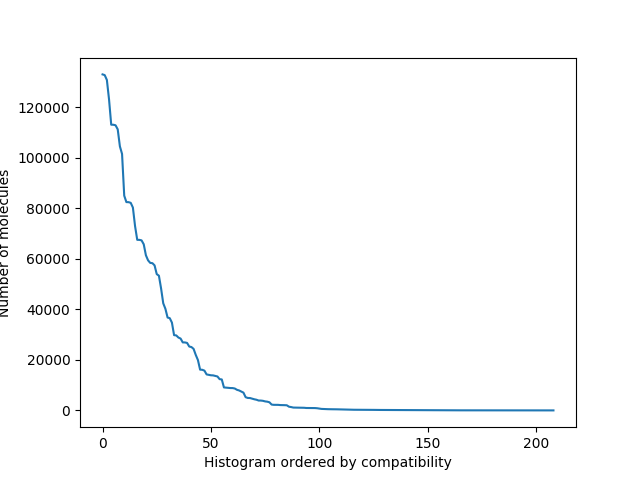}
\caption{dataset QM9.}
\end{subfigure}
\begin{subfigure}[t]{0.90\linewidth}
\centering
\includegraphics[width=\linewidth]{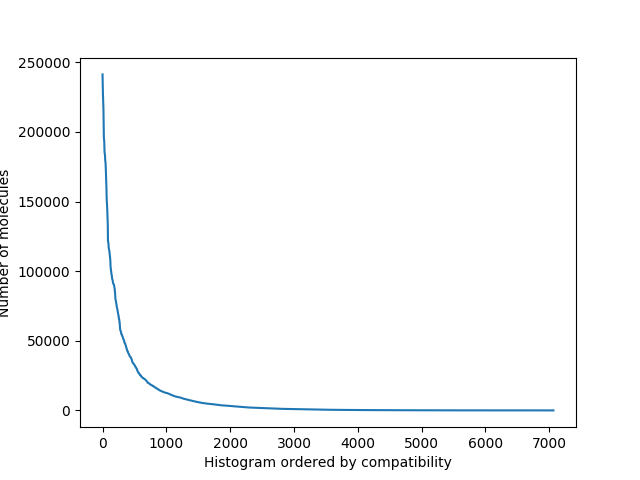}
\caption{dataset ZINC.}
\end{subfigure}
\caption{For each dataset, in the x-axis is reported the unique set of histograms, while, by focusing a precise histogram, on the y-axis there is the number of molecules (histograms) with which the histogram is compatible.}

\label{fig:histograms_distributions}
\end{figure}

Figure~\ref{fig:histograms_distributions} shows that, for both the QM9 and ZINC datasets, 
the histograms compatibility distributions are long tail distributions. Moreover, it is possible to see that very often the molecules have the same histogram of valences i.e. the unique set of the histogram of valences is formed by less then 300 histograms for the QM9 dataset and less than 8000 for the ZINC dataset.

Following the indications of~\cite{rigoni2020assessment} we considered the following metrics:
\begin{itemize}
    \item \emph{Reconstruction} that, given an input molecule and a set of generated molecules, computes the percentage of generated molecules that are equal to the one in input;
    \item \emph{Validity} that, given a set of generated molecules, represents the percentage of them that is valid, i.e. that represent actual molecules;
    \item \emph{Novelty} that represents the percentage of  generated molecules not in the training set;
    \item \emph{Uniqueness} that represents (in percentage) the ability of the model to generate different molecules in output, and is computed as the size of the unique set of valid generated molecules divided the total number of valid generated molecules;
    \item \emph{Diversity} that measures how much the generated molecules are different from those in the training set (comparing their substructures). This is a heuristic that uses randomly selected substructures present in the molecules.
    \item \emph{Natural Product (NP)} which indicates how much the generated molecules structural space is similar to the one covered by natural products~\cite{ertl2008natural};
    \item \emph{Solubility (Sol.)} which indicates how much a molecule is soluble in water, an important property for drugs;
    \item \emph{Synthetic Accessibility Score (SAS)} which represents how easy~(0) or difficult~(100) it is to synthesize a molecule;
    \item \emph{Quantitative Estimation Drug-likeness (QED)} which indicates in percentage how likely it is that the molecule is a good candidate to become a drug.
\end{itemize}

The \emph{Reconstruction} metric is calculated on 5000 test set molecules encoded 20 times in their latent space probability distributions and decoded one time in a molecule.
This process was chosen because both the encoder \p{e}{z|x,\theta{e}} and the decoder \p{d}{x|z,\theta{d}} contain a probabilistic component and in this way we estimate the model's ability to reconstruct the molecule considering both factors. So we treat the reconstruction joint probability of the molecule \x{} in input as:
\begin{displaymath}
\p{m}{x, z|x,\theta{e},\theta{d}} = \p{d}{x|z,\theta{d}} \cdot \p{e}{z|x,\theta{e}}
\end{displaymath}
where \(\theta{e}\) and \(\theta{d}\) are the neural networks parameters of the considered model.

Since we are interested in the generation of new molecules, the other metrics are computed using a different process that consists of directly sampling 20,000 points from the standard normal distribution and decoding each point only once.

\subsection{Reconstruction and Generation of new Molecules}

Table~\ref{tab:results} reports the average and standard deviation of the results obtained by the models on both the QM9 and ZINC datasets.
The last line of each table reports the properties scores obtained from the molecules in the datasets, while the last column of each table reports the number of epochs required for the model training. 
Note that, we used the same number of epochs as the CGVAE model.
As reported in section~\ref{subsec:data_metrics}, all the results regarding the QM9 dataset are different from the results reported in~\cite{rigoni2020assessment} due to the different data split used for the training and evaluation of the models, while the results regarding the dataset ZINC are the same.
However, for each dataset, all the models use the same split of data and the same procedure for generating the molecules and to perform the reconstruction task.
In particular, using a test set in which the molecules are more similar to those in the training set, the models \emph{reconstruction} ability tends to increase. 
In fact, in all models except Junction Tree VAE,
the \emph{reconstruction} performances are improved compared to~\cite{rigoni2020assessment}.

We can see from the table that our model improves the \emph{reconstruction} performance over CGVAE on both the datasets. 
In particular, the \emph{reconstruction} increases by $30.91\%$ on the QM9 dataset and by $21.82\%$ on the ZINC dataset.
Models based on the SMILES molecule representation, i.e. Character VAE, Grammar VAE and Syntax Directed VAE, usually present better \emph{reconstruction} values than models based on the molecule graph representation, but have problems to generate valid molecules, i.e. \emph{validity} values are low.
In fact, comparing our model \emph{reconstruction} values with the other models results, we can see that in the QM9 dataset, CCGVAE is the best among the models that deal directly with the molecular graphs, i.e. Graph VAE, Regularized GVAE, Junction Tree VAE, CGVAE.
Considering the ZINC dataset and only the models that deal directly with the molecular graphs, the Junction Tree VAE model presents the best reconstruction value. 
This is due to the fact that JTVAE uses common substructures to build the final molecule, so when reconstructing very complex molecules as those in ZINC, it is easier to use substructures than to sample every single atom. 
However, due to the use of substructures, JTVAE presents a very low value on the \emph{diversity} metric.
Overall, considering all the metrics (\emph{reconstruction}, \emph{validity}, \emph{novelty}, \emph{uniqueness} and \emph{diversity}), we can see that also in this case our model shows improved performances compared to CGVAE, trading off higher \emph{reconstruction} and \emph{diversity} with a slight decrease in \emph{uniqueness}.
If we consider the molecules' properties, our model presents better values regarding the \emph{NP} and the \emph{Sol.} metrics than CGVAE, but in the same way, our model presents worse values on the \emph{SAS} and the \emph{QED} metrics.
Note that, if we consider the \emph{NP} metric, in the QM9 dataset our model shows the highest values among the models results.
MolGAN, that is the only model based on the generative adversarial approach and applicable only to the QM9 dataset for computational reasons, presents lower results if compared with our model.
Overall, our model improves CGVAE in both the datasets, especially the use of the histogram of valences improves the performance on the \emph{reconstruction} task.
\begin{table*}[htbp]
\scriptsize
\centering
\begin{tabular}{| l | c | c | c | c | c " c | c | c | c " c |}
\hline
\dcrh{}\textbf{Model trained on QM9}      &$\uparrow$\textbf{\%Rec.}    &$\uparrow$\textbf{\%Val.}&\textbf{$\uparrow$\%Nov.}&\textbf{$\uparrow$\%Uniq.}   &\textbf{$\uparrow$\%Div.} &$\uparrow$\textbf{\%NP}  &$\uparrow$\textbf{\%Sol.} &$\downarrow$\textbf{\%SAS} &$\uparrow$\textbf{\%QED} &\textbf{N.Epochs}    \\
\hline
\mr{Character VAE}          &49.89          &5.86           &92.24          &\mr{94.80}         &91.31              &88.79          &\dcb{}46.55    &29.10          &30.02           &\mr{100}   \\
                            &\(\pm\)50.00   &\(\pm\)23.50   &\(\pm\)26.75   &                   &\(\pm\)18.93       &\(\pm\)11.74   &\(\pm\)32.71   &\(\pm\)28.52   &\(\pm\)19.55    &           \\
\hline
\mr{Grammar VAE}            &86.17          &12.59           &83.96         &\mr{59.27}         &98.72              &83.34          &35.85          &52.3          &35.18            &\mr{100}   \\
                            &\(\pm\)34.52   &\(\pm\)33.18   &\(\pm\)36.70   &                   &\(\pm\)6.67        &\(\pm\)15.45   &\(\pm\)19.46   &\(\pm\)31.63   &\(\pm\)11.53    &           \\
\hline
\mr{Syntax Directed VAE}    &\dcb{}97.54    &16.00          &\dcb{}100.00   &\dcb{}             &\dcb{}99.59        &88.89          &26.2           &14.65          &31.37           &\mr{500}   \\
                            &\(\pm\)16.00   &\(\pm\)36.66   &\(\pm\)0       &\dcb{}\mrm{100.00} &\(\pm\)1.14        &\(\pm\)10.64   &\(\pm\)22.26   &\(\pm\)35.15   &\(\pm\)11.18    &           \\
\hline
\mr{Graph VAE*}             &13.58          &80.06          &45.59          &\mr{88.09}         &66.18              &94.71          &35.92          &29.72          &48.25           &\mr{200}   \\
                            &\(\pm\)34.26   &\(\pm\)32.43   &\(\pm\)49.81   &                   &\(\pm\)28.01       &\(\pm\)10.82   &\(\pm\)13.49   &\(\pm\)28.27   &\(\pm\)9.53     &           \\
\hline
\mr{Regularized GVAE*}      &7.29           &91.77          &49.84          &\mr{77.13}         &68.65              &95.77          &39.38          &30.58          &\dcb{}48.79     &\mr{150}   \\
                            &\(\pm\)26.00   &\(\pm\)27.48   &\(\pm\)50.00   &                   &\(\pm\)25.63       &\(\pm\)9.26    &\(\pm\)14.52   &\(\pm\)24.69   &\(\pm\)7.83     &           \\
\hline
\mr{Junction Tree VAE}      &23.65          &99.92          &87.73          &\mr{89.48}         &60.85              &90.77          &27.25          &19.62          &46.89           &\mr{10}    \\
                            &\(\pm\)42.49   &\(\pm\)2.74    &\(\pm\)32.81   &                   &\(\pm\)29.49       &\(\pm\)16.00   &\(\pm\)13.17   &\(\pm\)21.18   &\(\pm\)7.73     &           \\
\hline
\mr{MolGAN}                 &               &85.11          &56.94          &\mr{17.68}         &62.29              &94.06          &34.63          &33.36          &48.44           &\mr{300}   \\
                            &\mrm{NA}       &\(\pm\)35.60   &\(\pm\)49.52   &                   &\(\pm\)34.70       &\(\pm\)35.22   &\(\pm\)17.52   &\(\pm\)23.90   &\(\pm\)18.66    &           \\
\hline
\mr{CGVAE*}                 &24.47          &\dcb{}100.00   &92.84          &\mr{98.34}         &76.12              &93.80          &28.62          &\dcb{}10.28    &47.91           &\mr{10}    \\
                            &\(\pm\)27.94   &\(\pm\)0.00    &\(\pm\)19.05   &                   &\(\pm\)22.64       &\(\pm\)5.62   &\(\pm\)12.38    &\(\pm\)16.11    &\(\pm\)7.04    &           \\
\hline
\textbf{\mr{CCGVAE (ours)}} &55.38          &\dcb{}100.00   &88.51          &\mr{93.18}         &79.16              &\dcb{}96.13    &35.58          &17.08          &46.62           &\mr{10}    \\
                            &\(\pm\)49.71   &\(\pm\)0.00    &\(\pm\)31.89   &                   &\(\pm\)22.02       &\(\pm\)8.64    &\(\pm\)11.91    &\(\pm\)22.96    &\(\pm\)7.51    &           \\
\hline
\hline
\dcrf{}                                 &\multicolumn{4}{c}{}                                   &                   & 88.52         & 27.91         &21.86          &46.12           &           \\
\dcrf{}\mrm{QM9 Properties' Scores}     &\multicolumn{4}{c}{}                                   &                   &\(\pm\)17.75   &\(\pm\)13.76   &\(\pm\)22.88   &\(\pm\)7.76     &           \\
\hline
\end{tabular}
\\\vspace*{0.1cm}
\begin{tabular}{| l | c | c | c | c | c " c | c | c | c " c|}
\hline
\dcrh{}\textbf{Model trained on ZINC}      &$\uparrow$\textbf{\%Rec.}    &$\uparrow$\textbf{\%Val.}&$\uparrow$\textbf{\%Nov.}    &$\uparrow$\textbf{\%Uniq.}   &$\uparrow$\textbf{\%Div.} &$\uparrow$\textbf{\%NP}  &$\uparrow$\textbf{\%Sol.}&$\downarrow$\textbf{\%SAS} &$\uparrow$\textbf{\%QED} &\textbf{N.Epochs}    \\
\hline
\mr{Character VAE*}         &25.28              &0.93           &\dcb{}100.00   &\mr{91.40}         &98.19          &80.82          &29.60          &31.11              &38.70          &\mr{100}   \\
                            &\(\pm\)43.46       &\(\pm\)9.60    &\(\pm\)0       &                   &\(\pm\)7.02    &\(\pm\)12.83   &\(\pm\)17.60   &\(\pm\)30.14       &\(\pm\)10.63   &           \\
\hline  
\mr{Grammar VAE*}           &55.82              &5.06           &\dcb{}100.00   &\mr{94.64}         &\dcb{}99.21    &80.99          &50.24          &26.75              &25.42          &\mr{100}   \\
                            &\(\pm\)49.66       &\(\pm\)22.99   &\(\pm\)0       &                   &\(\pm\)4.47    &\(\pm\)11.40   &\(\pm\)33.65   &\(\pm\)33.14       &\(\pm\)14.91   &           \\
\hline 
\mr{Syntax Directed VAE*}   &\dcb{}77.38        &19.00          &\dcb{}100.00   &\dcb{}             &93.56          &77.84          &55.94          &\dcb{}14.46        &39.45          &\mr{500}   \\
                            &\(\pm\)41.84       &\(\pm\)39.23   &\(\pm\)0       &\dcb{}\mrm{100.00} &\(\pm\)18.50   &\(\pm\)19.76   &\(\pm\)27.51   &\(\pm\)24.14       &\(\pm\)20.98   &           \\
\hline  
\mr{Graph VAE}              &0.27               &62.63          &\dcb{}100.00   &\mr{99.99}         &71.49          &90.68          &80.79          &28.07              &45.96          &\mr{400}   \\
                            &\(\pm\)4.58        &\(\pm\)48.38   &\(\pm\)0       &                   &\(\pm\)25.36   &\(\pm\)11.71   &\(\pm\)17.33   &\(\pm\)20.14       &\(\pm\)18.69   &           \\
\hline  
\mr{Regularized GVAE}       &0.01               &86.47          &\dcb{}100.00   &\mr{90.33}         &97.88          &\dcb{}95.88    &\dcb{}94.42    &44.64              &34.41          &\mr{300}   \\
                            &\(\pm\)0.77        &\(\pm\)34.21   &\(\pm\)0       &                   &\(\pm\)6.96    &\(\pm\)6.84    &\(\pm\)9.61    &\(\pm\)25.14       &\(\pm\)13.26   &           \\
\hline 
\mr{Junction Tree VAE*}     &50.23              &99.59          &99.98          &\mr{99.75}         &32.96          &52.20          &48.06          &44.74              &\dcb{}75.05    &\mr{10}    \\
                            &\(\pm\)50.00       &\(\pm\)6.35    &\(\pm\)1.23    &                   &\(\pm\)21.78   &\(\pm\)17.12   &\(\pm\)18.48   &\(\pm\)24.39       &\(\pm\)13.40   &           \\
\hline
\mr{CGVAE*}                 &0.35               &\dcb{}100.00   &\dcb{}100.00   &\mr{99.92}         &65.98          &81.38          &57.76          &16.25              &65.14          &\mr{3}     \\
                            &\(\pm\)5.91        &\(\pm\)0       &\(\pm\)0       &                   &\(\pm\)22.78   &\(\pm\)15.98   &\(\pm\)20.04   &\(\pm\)21.63       &\(\pm\)16.39   &           \\
\hline
\textbf{\mr{CCGVAE (ours)}} &22.17               &\dcb{}100.00   &\dcb{}100.00   &\mr{92.80}         &80.00          &94.28          &63.54          &19.95              &52.41          &\mr{3}     \\
                            &\(\pm\)41.54        &\(\pm\)0       &\(\pm\)0       &                   &\(\pm\)16.65   &\(\pm\)10.08   &\(\pm\)20.11   &\(\pm\)23.10       &\(\pm\)16.52   &           \\
\hline
\hline
\dcrf{}                     &\multicolumn{4}{c}{}                                                   &               & 42.08         & 56.11         & 55.95             & 73.18         &           \\
\dcrf{}\mrm{ZINC Properties' Scores} &\multicolumn{4}{c}{}                                          &               &\(\pm\)18.37   &\(\pm\)17.44   &\(\pm\)22.90       &\(\pm\)13.86   &           \\
\hline
\end{tabular}
\caption{Results obtained on the QM9 and ZINC data-sets. For each data-set: the symbol '*' denotes models where we used values for the parameters tuned by the authors for that data-set; entries with blue background highlight the best score obtained for each metric; up and down arrows in front of metrics name denote whether the metric should be maximized ($\uparrow$) or minimized ($\downarrow$). Average and standard deviation (where applicable) computed on the generated molecules are reported. Property scores for each data-set are reported as well.}
\label{tab:results}
\end{table*}

%

\section{Conclusions}
\label{sec:conclusions}
We have proposed Conditional Constrained Graph Variational Autoencoder (CCGVAE)
that, starting from a state-of-the-art model, uses the histogram of valences key-idea to guide the generation of the molecules, improving the performances of the base model in different performance metrics.

Future work will analyze and improve the computational time required for training the model.
Moreover, we will collaborate with chemists in order to incorporate more background knowledge in the  molecule generation process.

\section*{Acknowledgments}
The authors acknowledge the HPC resources of the Department of Mathematics, University of Padua, made available for conducting the research reported in this paper.

\bibliographystyle{IEEEtranN}
\bibliography{ref.bib}

\end{document}